\documentclass[a4paper,fleqn]{cas-dc}

\usepackage[numbers,sort&compress]{natbib}
\usepackage{amsmath}
\newdefinition{definition}{Definition}
\newdefinition{observation}{Observation}

\newcommand{\Ddel}{\ensuremath{\Delta_{\mathrm{del}}}}
\newcommand{\deldelS}{\ensuremath{\Delta_{\mathrm{del}}(S)}}
\newcommand{\deldelR}{\ensuremath{\Delta_{\mathrm{del}}(R)}}

\begin{document}

\let\WriteBookmarks\relax

\title[mode=title]{Deletion Sensitivity Is a Property of the Model: A
Three-Condition Measurement over 72 Valid Experimental Cells}
\shorttitle{Deletion Sensitivity Is a Property of the Model}
\shortauthors{Y.Li et al.}

\author[1]{Yiran Li}
\fnmark[1]
\author[1]{Hanzhang Jia}
\fnmark[1]
\cormark[1]
\ead{jhzaliyy@outlook.com}
\author[1]{Liheng Zeng}
\author[2]{Zhiyu Yang}
\author[3]{Yi Gao}
\cortext[1]{Corresponding author}
\fntext[1]{These authors contributed equally to this work.}
\affiliation[1]{organization={University of Sussex}, city={Brighton}, country={United Kingdom}}
\affiliation[2]{organization={Chang'an University}, city={Xi'an}, country={China}}
\affiliation[3]{organization={Shanghai Shuyuan Information Technology Co., Ltd.}, city={Shanghai}, country={China}}

\begin{abstract}
Reasoning models write a trace at every turn of a multi-turn
tool session, and later turns read it. Upstream providers hold
different retention constraints, so a request forwarded across issuers
arrives with its history deleted, and its task cost remains unmeasured.
This paper reports a direct, blind, multi-seed measurement of full
history deletion in multi-turn tool calling, through three rendering
conditions, a calibrated set of 72 valid experimental cells,
a multi-seed protocol, and the official checker.
The same deletion separates the models by direction and by magnitude,
one class losing tens of points, another
gaining, and a third unchanged, with the endpoint ordering reproduced
across seeds and task subsets modulating the magnitude and, near zero,
the sign. A re-dosed model preserves its success across
the full deletion range, so the reading belongs to the model rather
than the dose.
Deletion sensitivity is a model-level property under an operational
definition, measured under a fixed task family, scorer, and protocol,
and a retrained model is a new model that a two-cell screen, one
retention run and one deletion run, re-reads for the cost of two runs.
Deletion sensitivity is thus a measurable property of the model, and the
reading must enter routing as prior information rather than persist as
an invisible default cost.
\end{abstract}

\begin{keywords}
Chain-of-thought \sep Large language models \sep Multi-turn tool calling \sep Model evaluation \sep LLM serving
\end{keywords}

\maketitle

\section{Introduction}
\label{sec:intro}

Reasoning models now carry multi-turn tool calling. Within one session a
model issues calls, reads results, and plans the next step, and at every
turn it first writes a reasoning trace and then acts, while the next turn
reads the full history of traces and actions
\cite{wei2022cot,deepseek2025r1}. Across 94k ShareGPT conversations, 73
percent of user interactions span multiple turns \cite{wang2024mint}. The
trace is where the model keeps its plan, its intermediate results, and its
record of what it has already tried. The serving side, however, strips
the trace. Upstream providers hold provider-specific retention constraints
over the trace \cite{gemini2025,anthropic2025,openai2025,deepseekdocs2025},
a request forwarded across issuers arrives with its history deleted, and
this stripping is the common conservative rendering of the aggregation
ecosystem under which the serving condition of this paper is measured.
What that practice costs the task remains unmeasured.

Three gaps organize the question. What gets deleted is the reasoning
trace, and the actions and their results stay in the request. Prior work
on the value of the trace operates at the prompt layer of a single turn
\cite{wei2022cot,kojima2022zero,wang2023selfconsistency} or on its
faithfulness \cite{turpin2023,lanham2023,chen2025reasoning}, and the
accumulated history that a session carries across turns lies outside
those experiments. Where the deletion is visible has stayed unisolated,
since existing multi-turn evaluations hold the full history as the
default state of the session and keep deletion outside their variable
sets \cite{yan2024bfcl,kim2025taubench,wang2024mint,qin2024toolllm}.
How to measure it fairly is unsettled, since serving-side work reports
tokens saved and the tasks-lost column of its ledger stays empty
\cite{mu2023gist,jiang2023llmlingua,pan2024llmlingua2}. The measurement
this paper reports deletes the full reasoning history as the controlled
operation, grades the outcome with the official task checker, and repeats
it across seeds.

This paper measures the cost directly. The measurement renders each task
under
three conditions, retention, cross-turn deletion, and mid-session
deletion (Section~\ref{sec:apparatus}), scores every session with the
official BFCL checker blind to the condition, and reports the dual
indicator \deldelS{} and \deldelR{} with bootstrap intervals. The
calibrated dataset comprises 72 valid experimental cells spanning the
model configurations and task pools reported below, and one model adds a
deletion dose from 0 to 100 percent of its history. Table~\ref{tbl:compare} positions the measurement against the
surveyed lines of Section~\ref{sec:related}.

Under the fixed protocol, deletion sensitivity reads as a property of
the model version rather than a consequence of parameter count, as the
measurements of Section~\ref{sec:results} show. Every routing layer that forwards
traffic across issuer boundaries is therefore running an unquantified
cost. That cost is prior information for the routing decision, not a
default the deployment inherits, and this paper supplies the reading.
Deletion also leaves behavioral signatures that the final success score
cannot identify, and the repetition channel reads them for free.

This paper makes three contributions.
\begin{enumerate}
\item A direct, blind, multi-seed measurement of full
reasoning-history deletion, reported as the dual indicator \deldelS{} and
\deldelR{} with intervals, the reading a routing decision reads before
its first deletion.
\item The models separate by direction and by magnitude under one
deletion, from $-3.9$ to $+51.3$ points across the calibrated cells, with both
ends excluding zero, and neither parameter size nor deletion dose
explains the outcome, the 32B class alone spanning 51.5 points. The
task subset modulates the magnitude and flips the sign only where the
pooled effect is near zero.
\item Three behavioral signatures, compensation, looping, and quiet
failure, identified from actual call streams and not identifiable from the
final success score alone, together with a low-delta reference and a
two-cell screen, run the retention cell
and the deletion cell and read \deldelS{}, that turns the cost from an
invisible default into prior information read before the first deletion.
\end{enumerate}

\section{Related work}
\label{sec:related}

Several established lines of work concern the reasoning history of a
multi-turn tool-calling session. The closest lines are surveyed here, and
the present measurement is distinguished from each. Existing work divides
by
what is done to the reasoning trace and at which layer. One line operates
on the trace at the prompt layer, inside a single turn. A second line
evaluates multi-turn sessions with the full history present and holds
retention as the fixed default. A third line manipulates the history on
the serving side, along a spectrum of selective compression. The survey
treats training-time phenomena last, since they concern the production
of the trace, a question upstream of the serving side.
Table~\ref{tbl:compare} compares the lines by the operations they perform
on the trace, the grading they apply, and the replication they run.

\subsection{The reasoning trace at the prompt layer}
\label{sec:rw-prompt}

Before the reasoning models, the trace lived inside a prompt.
Chain-of-thought prompting established that writing the reasoning down
lifts single-turn capability \cite{wei2022cot}, the zero-shot form
extended the lift beyond worked examples \cite{kojima2022zero},
and self-consistency strengthened the lift by sampling the trace and
voting on the answer \cite{wang2023selfconsistency}. Two surveys place
these methods inside the general landscape, the first mapping the
concepts and methodologies of reasoning with foundation models
\cite{sun2025reasoning}, the second organizing the multi-step methods
that interleave generation and verification \cite{plaat2024multistep}.
However, with the emergence of the reasoning models, the trace moved
from a prompt technique to a first-class output. The o1 system learns to
reason at training time and emits long traces as its working medium
\cite{openai2024o1}, DeepSeek-R1 arrives at the same behavior through
reinforcement learning \cite{deepseek2025r1}, and the Qwen3 family ships
hybrid thinking modes across the size ladder this study measures
\cite{qwen3_2025}. The trace grew with the transition, a reasoning model
writes hundreds to thousands of trace tokens per turn, and the history
of a multi-turn session now carries more trace than answer.

A second wave of this family measures whether the trace reports the
computation that produced the answer, and finds a recurring gap between
the two. Unfaithful explanations appear under chain-of-thought prompting
\cite{turpin2023}, truncating the trace leaves the answer intact often
enough to question the role of the visible steps \cite{lanham2023}, and
the reasoning models inherit the gap between their stated and their
actual reasoning \cite{chen2025reasoning}. Two traits separate this
family from the present work. First, the unit of operation is the prompt
of one turn, and the accumulated history that a session carries across
turns stays outside those experiments. Second, the question asked of the
trace is whether it helps or whether it is faithful, and these
experiments pose a different question from the cost of deleting the
accumulated history. The faithfulness line opens the possibility that
the trace costs little to remove, and settling that possibility at the
history layer is exactly the question this paper measures.

\subsection{Evaluations that keep the history}
\label{sec:rw-eval}

The evaluations of this family place a model in a multi-turn session and
grade the outcome, and the history arrives intact at every turn. The
family grew out of the tool-use line. Toolformer taught a model to call
tools mid-sentence and evaluated the calls in isolation
\cite{schick2023toolformer}, Berkeley Function Calling turned the call
format into a leaderboard and shipped the official checker that grades
terminal state, call order, and no-call turns \cite{yan2024bfcl,bfcl2025},
ToolLLM extended the coverage to thousands of APIs in single-turn and
chained forms \cite{qin2024toolllm}, and the tool-learning survey now
maps the whole curriculum from tool selection to execution
\cite{qu2024tool}. The multi-turn line follows. ReAct interleaved the
trace with the action and is the direct ancestor of the trace--answer
split this study operates on \cite{yao2023react}, $\tau$-bench runs a
model against a user simulator across turns \cite{kim2025taubench}, MINT
interleaves tools with language feedback \cite{wang2024mint}, and two
surveys organize the ground, the first a general evaluation survey for
large language models \cite{chang2025evalsurvey}, the second a dedicated
survey of agent evaluation that reaches tool use and memory
\cite{yehudai2026agents}.

A second line of this family measures how the session itself fares as
the history grows, with the history present at every turn. Models lose
the thread as turns accumulate even with the full history intact
\cite{laban2025compress}, position effects compound the loss inside long
contexts \cite{liu2024lost}, and model-graded evaluation, the method
many of these leaderboards lean on, is itself a surveyed instrument with
known biases \cite{zheng2023judge}. In all of these, the full history is
the default state of the session, every design delivers the session
intact and grades it, and the stripped request, the state production
traffic runs in, stays outside the design.

\subsection{Serving-side manipulation of the history}
\label{sec:rw-serving}

The systems of this family rewrite the history between turns, on the
serving side, and they report what is saved. The rewriting line developed
its own curriculum. Gist tokens compress prompts into learned tokens
that stand in for the original text at inference time \cite{mu2023gist},
LLMLingua prunes the prompt token by token under a small auxiliary
scorer \cite{jiang2023llmlingua}, LLMLingua-2 trains the pruning decision
through data distillation and buys a further speedup
\cite{pan2024llmlingua2}, and the compression field is now surveyed as a
spectrum from token pruning to learned summaries
\cite{li2025promptsurvey}. Below the tokens, the serving stack has its
own economics, PagedAttention organized the KV cache into pages and made
the memory budget the unit of scheduling \cite{xiao2023vllm}, and the KV
side now carries its own survey of compression and eviction
\cite{jiang2025kvsurvey}. What this literature reports is tokens saved
and latency gained, and the task cost of the rewrite is outside its
tables.

\paragraph{The provider boundary.}
Four provider policies govern the trace, and Table~\ref{tbl:providers}
compares them, one field per provider, and the four differ in how the
field is honoured. Provider-specific retention constraints govern
cross-issuer history, and a request forwarded across issuers arrives with
its history deleted in full
\cite{gemini2025,anthropic2025,openai2025,deepseekdocs2025}.
The stripping is the common conservative rendering of the aggregation
ecosystem, full deletion is what every issuer boundary admits, and the
practice is the reason the cost of the stripped request is a
measurement worth taking.

\paragraph{The closest measurements.}
Two recent works come closest to this one, and two facts separate each
from it. The first work lets agents forget their reasoning under a
selective scheme on a retention spectrum \cite{iclr2026compress}. The
second work measures when a model should think at all, scoring dependence
on visible reasoning \cite{race2026}. The first fact is the operation.
Compression and selective forgetting are design choices on a spectrum an
operator controls, while full deletion is the forced outcome at an
issuer boundary that every operator inherits, and the two are different
objects. The second fact
is the reading. Those works report retention ratios and dependence
scores, while this work reports the task cost of the complete deletion,
graded by the official checker, blind, across seeds and conditions, with
intervals. Prior serving-side work reports tokens saved. This
measurement reports tasks lost, holding the engine, the task pool, and the
scorer fixed, under three controlled conditions.

\subsection{Training-time phenomena}
\label{sec:rw-training}

A final line observes the trace at training and release time. Trace
collapse degrades the reasoning trace under training pressure
\cite{twist2026collapse}, and the reasoning profile of a model family
shifts across releases \cite{rodionov2026shift}. These phenomena concern
the production of the trace rather than its serving-side fate, and they
enter this study through the finding that deletion sensitivity is a
property of the specific model version, which the training-time line
makes plausible and the present measurements quantify.

\subsection{Key differences}
\label{sec:rw-keydiff}

Over the systems surveyed above, this work adds three things. It holds
the full deletion of the reasoning history as the controlled operation of
the experiment, across issuer boundaries, in the direction production
traffic actually flows. It grades the outcome of that operation with the
official task checker under a blind, multi-seed protocol. And it reports
the cost as a model-level property with intervals, which is the reading a
deployment can act on before the first deletion happens. The design,
the protocol, and the reading supply these three
(Sections~\ref{sec:setup}--\ref{sec:results}).

\begin{table*}[t]
\caption{How the four providers treat the reasoning history. The four
differ in how each honours the field, and cross-issuer history is the
condition under which full deletion is the common conservative
rendering.}\label{tbl:providers}
\begin{tabular}{@{}llll@{}}
\toprule
Provider & Trace visibility & Cross-turn default & Stance \\
\midrule
OpenAI & encrypted, bound to issuer & renders within lineage & recommends carrying \\
Anthropic & summary plain, body encrypted & strips old turns server-side & requires current loop \\
Gemini & encrypted thought signatures & skips old-turn validation & warns of degradation \\
DeepSeek & plain text & API ignores the field & advises discarding \\
\bottomrule
\end{tabular}
\end{table*}

\begin{table*}[t]
\caption{This measurement against prior evaluation and serving-side work.}\label{tbl:compare}
\begin{tabular}{@{}lcccc@{}}
\toprule
Work & Multi-turn & Deletion & Blind grading & Seeds \\
\midrule
Forget selectively \cite{iclr2026compress} & partial & selective & -- & -- \\
When to think \cite{race2026} & -- & dependence & -- & -- \\
Shift and collapse \cite{rodionov2026shift,twist2026collapse} & -- & training-time & -- & -- \\
BFCL \cite{yan2024bfcl} & yes & -- & yes & -- \\
Lost in multi-turn \cite{laban2025compress} & yes & -- & -- & -- \\
This work & yes & full, three cells & yes & yes \\
\bottomrule
\end{tabular}
\end{table*}

\section{Experimental design and setup}
\label{sec:setup}

\subsection{Task pool and models}
The task set is the four multi-turn subsets of BFCL v4, 794 tasks with 2
to 8 turns across base scenarios, long context, missing functions, and
missing parameters \cite{yan2024bfcl,bfcl2025}. Three properties fix the
choice. The tasks execute against a real tool environment, the sessions
are multi-turn by construction, and the leaderboard ships an official
checker, which Section~\ref{sec:grading} turns into the blind grader.
Five models run the full 794-task scale and three run the 200-task
scale, and the valid-session denominators ride with every table, so a
comparison reads its own pool rather than a pooled count.

Table~\ref{tbl:models} lists the model configurations represented in the calibrated dataset. The three Qwen3 models
form a size ladder inside one family at 8B, 14B, and 32B
\cite{qwen3_2025}, GLM-Z1-32B and Baichuan-M2-32B are two further 32B
models from other houses \cite{glmz1}, GLM-Z1-9B-0414 extends the GLM
house at 9B \cite{glmz1}, Ling-mini-2.0 is a further open reasoning
model at 16.8B, and R1-Distill-14B is a distillation
lineage \cite{deepseek2025r1}. Every model carries the complete grid,
three conditions at three seeds, and every table pools the seeds
within its cells.

\begin{table}[t]
\caption{Model configurations in the 72 experimental cells. Params are in billions. Valid-session totals pool three seeds and list retention then cross-turn deletion. Models on the 200-task scale use that task pool.}\label{tbl:models}
\begin{tabular}{@{}lrrr@{}}
\toprule
Model & Params & Tasks & Valid sessions \\
\midrule
Qwen3-8B & 8 & 794 & 2382/2380 \\
Qwen3-14B & 14 & 794 & 2375/2382 \\
Qwen3-32B & 32 & 794 & 2382/2382 \\
GLM-Z1-32B & 32 & 794 & 2302/2232 \\
GLM-Z1-9B-0414 & 9 & 200 & 567/591 \\
Baichuan-M2-32B & 32 & 200 & 596/573 \\
Ling-mini-2.0 & 16.8 & 200 & 569/567 \\
R1-Distill-14B & 14 & 794 & 1852/1892 \\
\bottomrule
\end{tabular}
\end{table}

\subsection{Three rendering conditions}
\label{sec:apparatus}
\begin{figure*}[pos=!t]
\centering
\includegraphics[width=\textwidth]{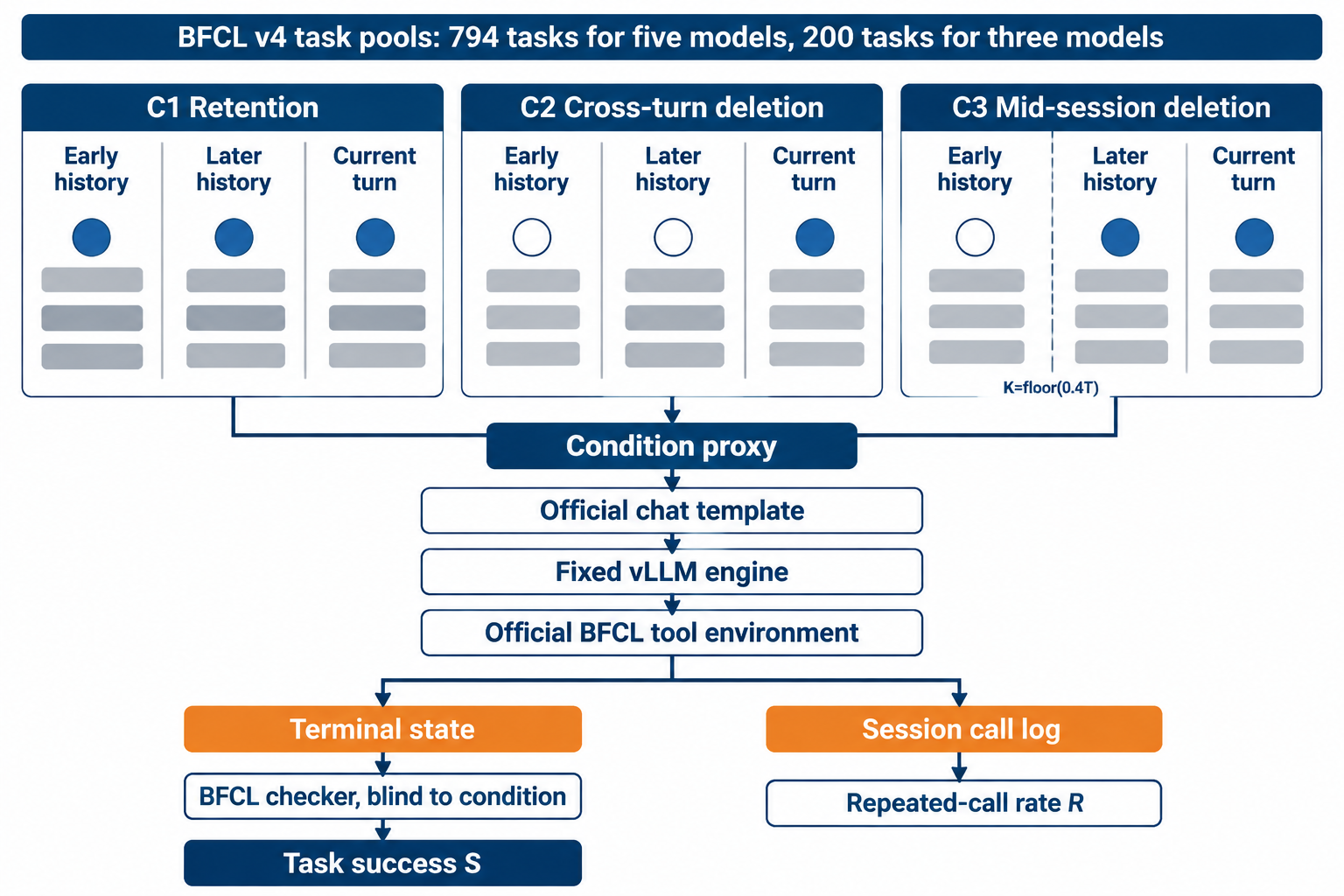}
\caption{The measurement design. Filled circles mark retained traces and open circles mark deleted traces in the early history, later history, and current turn. Gray bars retain actions and results. The condition proxy applies each rendering condition before the shared chat template, fixed engine, and official tool environment. The official BFCL checker scores terminal task state while blind to condition, and the call log supplies the repeated-call rate.}\label{fig:apparatus}
\end{figure*}

Let a session be $m=(u_1,r_1,\dots,u_T,r_T)$ with turn record
$r_t=(c_t,s_t)$, where $c_t$ is the reasoning trace of turn $t$ and $s_t$
its actions and answer. For a set $A$ of turns, the deletion operator is
\begin{equation}
D_A(m)_t=\begin{cases}(\bot,s_t), & t\in A,\\ (c_t,s_t), & t\notin A,\end{cases}
\label{eq:delop}
\end{equation}
and the three conditions render
\begin{align}
R_{C1}&=\mathrm{render}(m),\notag\\
R_{C2}&=\mathrm{render}(D_{1..T-1}(m)),\notag\\
R_{C3}&=\mathrm{render}(D_{1..K}(m)),\quad K=\lfloor 0.4T\rfloor.
\label{eq:conds}
\end{align}
The current tool loop, the burst of calls and results inside the current
turn, keeps its trace under every condition, because providers reject a
request that drops it \cite{anthropic2025,gemini2025}. Historical turns
are everything before the current turn, and deletion operates exactly
there. The model itself enters as a function, and the turn $t$ answer
follows
\begin{equation}
y_t=f_\theta\big(\mathrm{render}(D_A(m_{<t})),\,u_t\big),
\label{eq:gen}
\end{equation}
where $f_\theta$ is the engine-loaded model, $m_{<t}$ the message
prefix before turn $t$, $A$ the turn set the condition marks for
deletion, and $u_t$ the user turn. Equation~(\ref{eq:conds}) is the
end-of-session view of the same rendering, and the conditions change
$A$ alone, while $f_\theta$, the task environment, and the grader stay
fixed throughout. The mid-session condition switches inside the session
at $K=\lfloor 0.4T\rfloor$, deleting the reasoning of the first 40 percent
of turns and keeping the rest, which matches the deployed
implementation this study reproduces. Figure~\ref{fig:apparatus} shows
the three history renderings and the separate success and repetition
measurements. The
retention and cross-turn pair reads the total sensitivity of a model in
two cells, and the mid-session arm isolates the cost of the switch
itself.

\begin{figure*}[pos=!t]
\centering
\includegraphics[width=\textwidth]{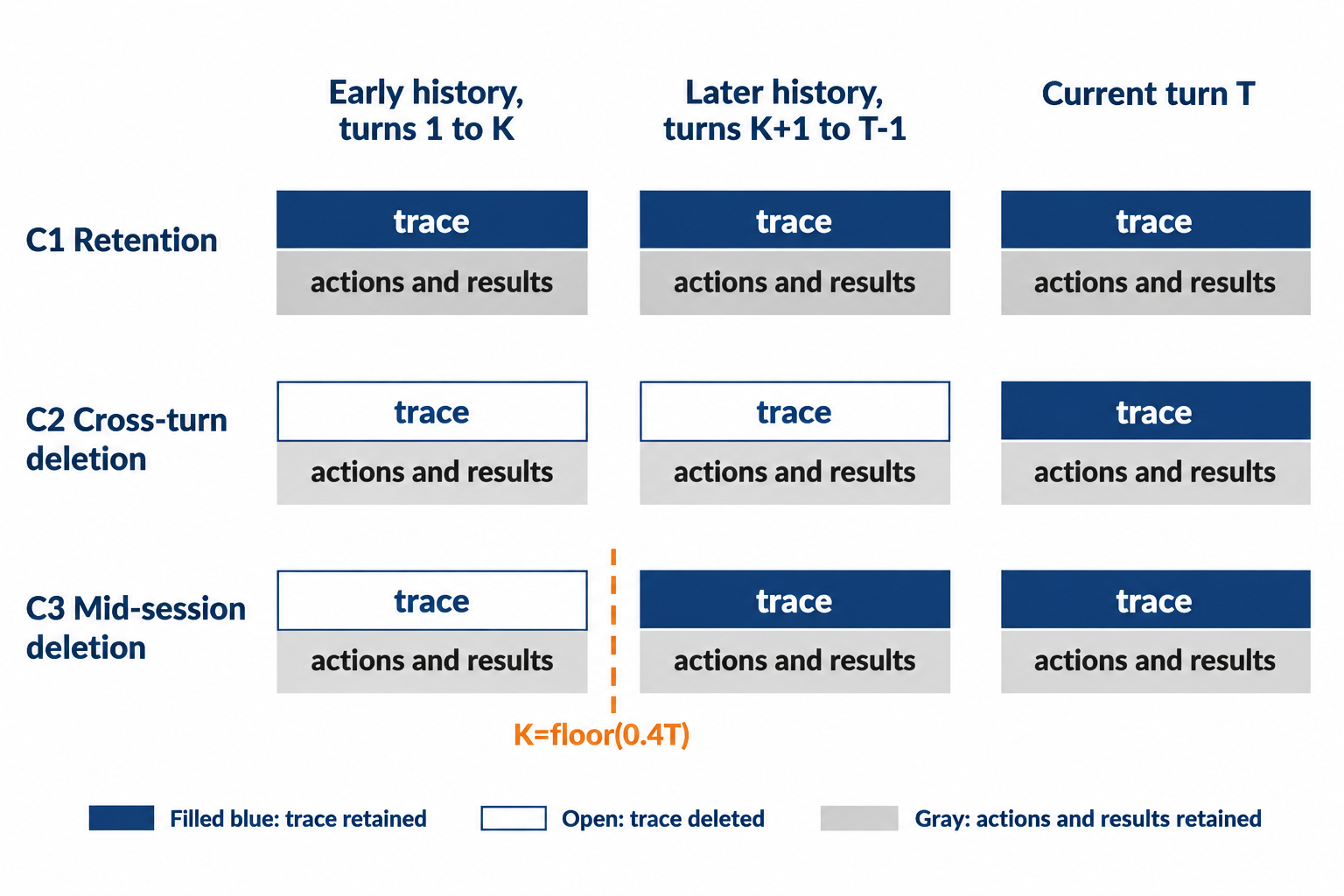}
\caption{Message-level construction of a session under retention, cross-turn deletion, and mid-session deletion. Filled blue blocks mark retained traces, open blocks mark deleted traces, and gray bars retain actions and results. The dashed boundary marks $K=\lfloor0.4T\rfloor$ in the mid-session condition, while the current turn retains its trace in every condition.}\label{fig:conditions}
\end{figure*}

Figure~\ref{fig:conditions} details how each condition renders traces,
actions, and results across the session.

\subsection{Deletion sensitivity metrics}
\begin{equation}
\Ddel(S)=S(C1)-S(C2),\qquad \Ddel(R)=R(C2)-R(C1),
\label{eq:delta}
\end{equation}
with $S$ in percent (Eq.~\ref{eq:score}). The two indicators carry
opposite signs on purpose, a sensitivity loss and a repetition rise both
read positive. Each interval comes from session-level bootstrap,
\begin{equation}
\mathrm{CI}_{95}=\big[q_{2.5\%}(\Delta^{*}),\,
q_{97.5\%}(\Delta^{*})\big],
\label{eq:ci}
\end{equation}
where $\Delta^{*}$ is the indicator, \deldelS{} or \deldelR{}, recomputed
on a resample of the valid sessions with replacement, $B=10{,}000$
resamples drawn at the level of the task, the resampling unit that pairs
the retention and deletion cells of one task and keeps the pairing inside
every scan, and $q_\alpha$ the empirical $\alpha$-quantile over the $B$
copies.

\begin{definition}[deletion sensitivity as a model property]
\label{def:property}
Under a fixed task family, scorer, and sampling protocol, deletion
sensitivity is an operational property of the model version, estimated by
\deldelS{} with its interval. The measured cells establish seed stability
for the reported responses and show that parameter size does not order
them. Dose invariance belongs only to the model directly dosed, while the
task subset modulates magnitude and can flip the sign when the pooled
effect is near zero. A new model version or task family therefore requires
a new two-cell reading, as Section~\ref{sec:ana-transfer} specifies.
\end{definition}

\subsection{Implementation, grading, and cleaning}
\label{sec:grading}
The session harness is a modified version of the open-source DeepSeek
Harness.\footnote{\url{https://github.com/deepseek-ai/deepseek-harness},
developer preview, version 0.1.0-rc.5.}
A harness drives each session turn by turn, the model answers through a
condition proxy at the message layer, the proxy re-renders the whole
request through the official chat template of the model so that the
rendered form of every surviving message is identical across conditions,
and a self-hosted vLLM engine \cite{xiao2023vllm} executes the request
against the official task environment. The engine holds each model at a
fixed configuration for all three conditions, tensor parallelism across
two devices, a 32K context window for the 32B models and 40K for the
Qwen3 ladder, temperature 0.6, and a generation budget that carries every
trace to its end. Every model runs three seeds per condition, and the
cleaning rule below applies to all cells alike.

Scoring follows the official BFCL form,
\begin{equation}
S(C)=\frac{100}{|Q|}\sum_{q\in Q}\mathbf{1}\!\left[\mathrm{checker}(q,
\mathrm{final\_state}(C,q))\right],
\label{eq:score}
\end{equation}
where $q$ is a task, $Q$ the task pool, $\mathrm{final\_state}(C,q)$ the
end state of the session under condition $C$, and $\mathrm{checker}$ the
official BFCL grader, which returns 1 when the session passes and 0
otherwise. The repetition rate carries its own numbered form,
\begin{equation}
R(C)=\frac{100}{|Q|}\sum_{q\in Q}\mathbf{1}\!\left[\mathrm{twice}(C,q)
\right],
\label{eq:rep}
\end{equation}
where $\mathrm{twice}(C,q)$ holds when the session of task $q$ under
condition $C$ issues at least one identical call twice. Dose arms
$d0$--$d100$ run on the one model dosed. Scoring is blind to the
condition label, and the checker inspects the terminal state, the
call order, and the no-call turns of every session. A session whose
failure cannot be attributed to the model or to the infrastructure
leaves the denominator, with transport, driver, and engine failures
recognized as infrastructure faults.

Table~\ref{tbl:main} reports bootstrap intervals, which remain the
reporting standard of the study, and the screening arithmetic of
Section~\ref{sec:deploy} carries its own closed form.

\subsection{Validity checks}
Three checks establish that the readings belong to the models under
measurement, and each check closes one artifact route. First, the scorer
reproduces
cell by cell the deployed scoring of this project, all nine Qwen3-8B
cells score identically under the local and the deployed scorer, 231,
286, 302, 258, 269, 304, 289, 274, and 298 valid tasks. Second, the
readings are stable across seeds, Baichuan-M2-32B reads $+53.0$,
$+47.2$, and $+53.8$ points on its three cross-turn seeds, every seed
large and positive. Third, the measurement separates sampling resolution
from signal. An earlier 40-task three-seed measurement of GLM-Z1-32B
read 53.3 against 43.3 percent, retention against cross-turn deletion
\cite{glmz1}, and the full-scale tables read the same contrast at $+1.3$
points with the interval $[-1.3, +3.9]$. The small-scale gap sits inside
the noise band that Eq.~(\ref{eq:se}) assigns to a cell of that size,
the full-scale interval is several times tighter, and the reading
stabilizes as the session count grows, which is the reporting standard
behind every table of this study.

\section{Results}
\label{sec:results}
\subsection{Sensitivity spans $-3.9$ to $+51.3$ points}
\label{sec:res51}

\begin{figure*}[pos=!t]
\centering
\includegraphics[width=\textwidth]{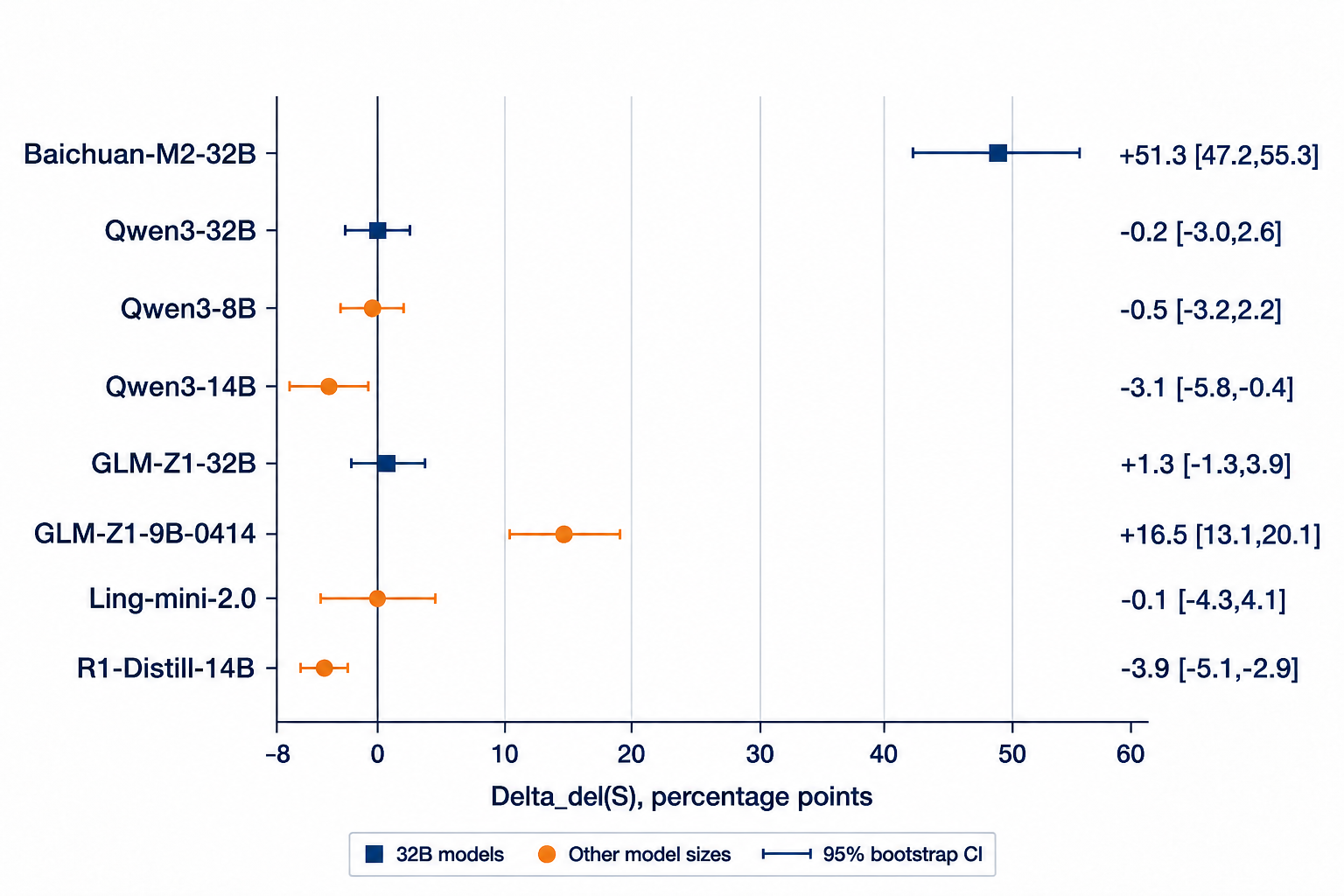}
\caption{Deletion sensitivity by model, ordered by retention success. Points mark estimates and horizontal bars show 95 percent bootstrap confidence intervals. Squares mark 32B models and circles mark other sizes.}\label{fig:forest}
\end{figure*}

The calibrated cells place the same deletion operation 55.2 points apart across model configurations.
Table~\ref{tbl:main} reports the main result and
Figure~\ref{fig:forest} orders the models by their retention baseline.
Retention minus cross-turn deletion runs from $-3.9$ points on
R1-Distill-14B, where deletion improves the model, to $+51.3$ points on
Baichuan-M2-32B, where deletion erases it, and both intervals exclude
zero. Deletion erases Baichuan-M2-32B, its cross-turn cells hold 3 valid
successes out of 573 valid sessions, it erases GLM-Z1-9B-0414 the same
way, 16 valid successes out of 591, and deletion improves
R1-Distill-14B by 3.9 points within a low absolute-success regime between
0.9 and 4.9 percent. GLM-Z1-32B holds its success under cross-turn deletion,
retention and cross-turn read 29.1 and 27.8 percent, Ling-mini-2.0 reads
zero difference at a 15 percent baseline, and the three Qwen3
models sit within 3.1 points of zero in either direction.

\begin{table*}[t]
\caption{Success rates in percent, with successful sessions over valid sessions in parentheses, pooled across three seeds. \deldelS{} is retention minus cross-turn deletion in percentage points with its 95 percent bootstrap interval. \deldelR{} is cross-turn deletion minus retention in repetition rate.}\label{tbl:main}
\begin{tabular*}{\textwidth}{@{\extracolsep{\fill}}lrrrrrr@{}}
\toprule
Model & Retention & Cross-turn & Mid-session & \deldelS{} & 95\% interval & \deldelR{} \\
\midrule
Baichuan-M2-32B & 51.8 (309/596) & 0.5 (3/573) & 14.7 (93/632) & $+51.3$ & $[+47.2, +55.3]$ & $+8.0$ \\
GLM-Z1-9B-0414 & 19.2 (109/567) & 2.7 (16/591) & 6.3 (37/590) & $+16.5$ & $[+13.1, +20.1]$ & $-7.2$ \\
GLM-Z1-32B & 29.1 (671/2302) & 27.8 (621/2232) & 23.8 (551/2319) & $+1.3$ & $[-1.3, +3.9]$ & $-0.1$ \\
Ling-mini-2.0 & 15.3 (87/569) & 15.3 (87/567) & 15.4 (87/565) & $-0.1$ & $[-4.3, +4.1]$ & $-3.8$ \\
Qwen3-8B & 34.4 (819/2382) & 34.9 (831/2380) & 36.2 (861/2381) & $-0.5$ & $[-3.2, +2.2]$ & $+1.7$ \\
Qwen3-14B & 34.4 (816/2375) & 37.4 (892/2382) & 35.4 (843/2382) & $-3.1$ & $[-5.8, -0.4]$ & $-0.7$ \\
Qwen3-32B & 40.3 (960/2382) & 40.5 (965/2382) & 39.2 (933/2382) & $-0.2$ & $[-3.0, +2.6]$ & $+1.4$ \\
R1-Distill-14B & 0.9 (17/1852) & 4.9 (92/1892) & 3.1 (59/1875) & $-3.9$ & $[-5.1, -2.9]$ & $+5.9$ \\
\bottomrule
\end{tabular*}
\end{table*}

Parameter class leaves the outcome open. The three 32B models receive
the same operation and read $-0.2$, $+1.3$, and $+51.3$ points, a
51.5-point spread inside one parameter class, and the GLM house carries
the contrast inside one lineage, its 32B reads $+1.3$ while its 9B
reads $+16.5$, while the two 14B models
sit at $-3.1$ and $-3.9$ and the 8B model sits at $-0.5$. A deployment
that trusted size as the predictor would delete the history of a 32B
model and lose half its success.

The outcomes admit three descriptive response groups. The small-response group
holds the three Qwen3 models, GLM-Z1-32B, and Ling-mini-2.0, where
deletion moves success by
at most 3.1 points in either direction. The recovering group holds
R1-Distill-14B, which gains 3.9 points within a low absolute-success
regime between 0.9 and 4.9 percent. The collapsing group holds Baichuan-M2-32B and
GLM-Z1-9B-0414, where deletion costs 16.5 to
51.3 points.

The endpoint responses are stable across seeds. Baichuan-M2-32B reads $+53.0$,
$+47.2$, and $+53.8$ points on its three cross-turn seeds, every seed
large and positive, GLM-Z1-9B-0414 reads $+17.6$, $+17.7$, and $+14.3$,
the nine Qwen3 cross-turn readings all sit within
4.8 points of zero, and the three GLM-Z1-32B cross-turn seeds read
$+2.5$, $-0.3$, and $+1.7$, so sampling perturbation
keeps the endpoint ordering in place.

\subsection{What the outcome tracks}
\label{sec:res52}

For the one model of the small-response group that we dose, recovery is
invariant from 0 to 100 percent deletion.

\begin{observation}[dose invariance for the dosed model]
\label{obs:dose}
For the one model dosed, the d0 and d100 endpoints differ by 1.5
points, and the interval of every dose arm's difference from the zero
arm crosses zero.
\end{observation}

\begin{figure*}[pos=!t]
\centering
\includegraphics[width=\textwidth]{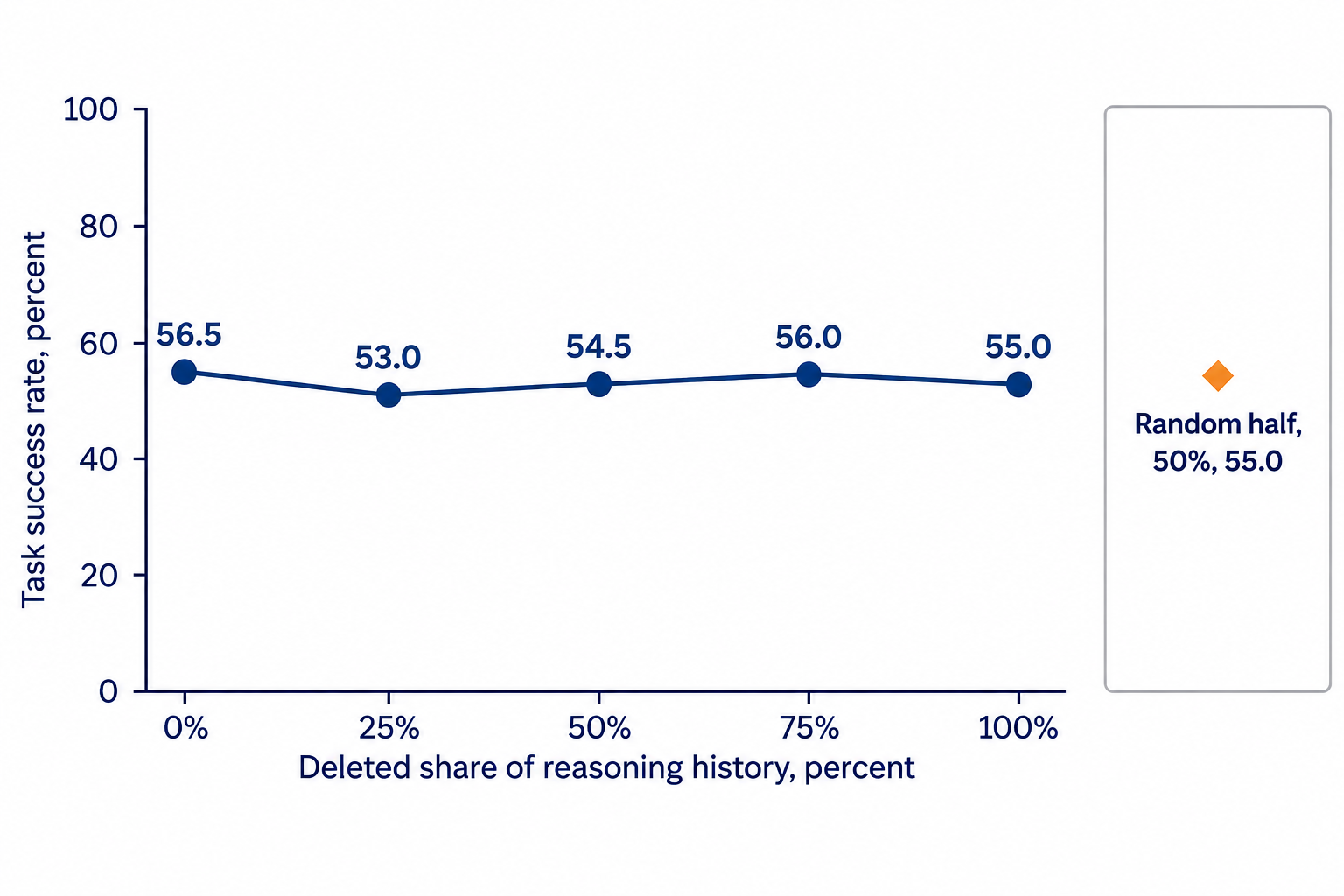}
\caption{Qwen3-32B task success across five ordered deletion doses. The separate diamond marks the random-half arm, and the vertical axis spans 0 to 100 percent. The d0 and d100 endpoints differ by 1.5 percentage points, and the six observed rates span 3.5 points.}\label{fig:dose}
\end{figure*}

The 32B Qwen3 model runs six arms on the
200-task subset, deleting 0, 25, 50, 75, and 100 percent of the history
plus a second 50 percent arm that deletes a different random half, and
success reads 56.5, 53.0, 54.5, 56.0, 55.0, and 55.0 percent. The d0
and d100 endpoints differ by 1.5 points, the interval of every dose
arm's difference from the zero arm crosses zero, and random-half
deletion matches block-half deletion at 54.5
against 55.0. Success is preserved at every dose, and whether the model
re-derives the deleted steps or carries the trace as idle cargo is the
mechanism question of Section~\ref{sec:ana62}. The dose claim therefore
belongs to the dosed model alone and does not enter the cross-model
predictor claim.
Figure~\ref{fig:dose} plots the five ordered deletion doses and marks the
random-half arm separately.

Task subsets modulate the magnitude, and the sign follows the model
wherever the pooled effect sits away from zero.
Table~\ref{tbl:subsets} decomposes \deldelS{} by subset. Baichuan-M2-32B
pays 64.7 points on base scenarios and 59.3 on long context, drops to
45.6 on missing parameters and 7.8 on missing functions, where the
absent tool bounds what any condition can save, and GLM-Z1-9B-0414
follows the same shape at a smaller scale, 22.5 on long context and 1.2
on missing functions. GLM-Z1-32B reads $-2.7$ on missing functions
against its pooled $+1.3$, the one cell whose interval excludes zero in
the direction opposite to its pooled sign, so the small-response group is the
group whose sign the subset moves. R1-Distill-14B keeps its improvement
on all four subsets, the near-zero Qwen3 models stay within 7.4
points of zero everywhere, and the subset mix therefore moves the
magnitude and leaves the classification of the endpoints alone.

\begin{table*}[t]
\caption{Subset-specific \deldelS{} in percentage points with 95 percent bootstrap intervals, for retention minus cross-turn deletion.}\label{tbl:subsets}
\footnotesize
\begin{tabular}{@{}lrrrr@{}}
\toprule
Model & Base scenarios & Long context & Missing functions & Missing parameters \\
\midrule
Baichuan-M2-32B & $+64.7$ $[+57.5, +70.9]$ & $+59.3$ $[+52.3, +65.8]$ & $+7.8$ $[+2.1, +15.2]$ & $+45.6$ $[+34.1, +55.9]$ \\
GLM-Z1-9B-0414 & $+16.4$ $[+10.3, +22.6]$ & $+22.5$ $[+16.2, +29.0]$ & $+1.2$ $[-4.0, +7.0]$ & $+18.2$ $[+8.3, +29.4]$ \\
GLM-Z1-32B & $+2.4$ $[-3.3, +8.1]$ & $+3.8$ $[-1.9, +9.5]$ & $-2.7$ $[-4.8, -0.8]$ & $+2.7$ $[-2.4, +7.7]$ \\
Ling-mini-2.0 & $-1.2$ $[-8.4, +6.0]$ & $-1.0$ $[-8.4, +6.4]$ & $+2.3$ $[-5.0, +10.0]$ & $+3.2$ $[-9.3, +15.4]$ \\
Qwen3-8B & $-2.2$ $[-7.9, +3.5]$ & $-0.8$ $[-6.5, +4.8]$ & $+2.0$ $[-0.7, +4.7]$ & $-1.1$ $[-6.5, +4.4]$ \\
Qwen3-14B & $-7.4$ $[-13.0, -1.7]$ & $-3.6$ $[-9.2, +2.1]$ & $+0.0$ $[-2.8, +2.8]$ & $-1.3$ $[-6.8, +4.2]$ \\
Qwen3-32B & $-0.8$ $[-6.5, +4.8]$ & $-3.9$ $[-9.5, +1.8]$ & $+0.0$ $[-2.7, +2.7]$ & $+3.8$ $[-1.8, +9.4]$ \\
R1-Distill-14B & $-4.9$ $[-7.8, -2.2]$ & $-6.9$ $[-10.1, -3.9]$ & $-3.0$ $[-4.8, -1.6]$ & $-1.5$ $[-3.1, -0.3]$ \\
\bottomrule
\end{tabular}
\end{table*}

\begin{figure*}[pos=!t]
\centering
\includegraphics[width=\textwidth]{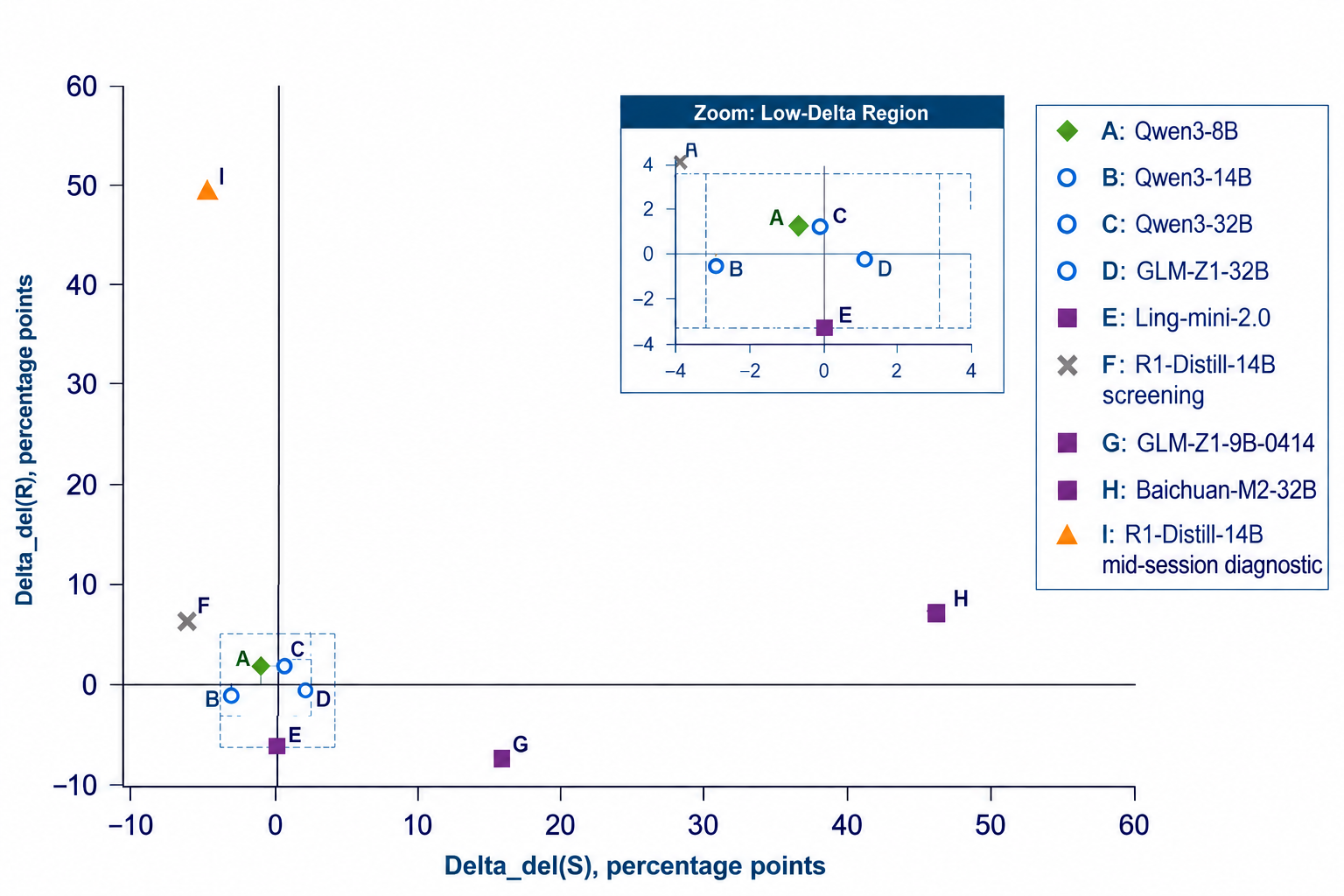}
\caption{The two axes show success and repetition changes in percentage points. The shaded rectangle marks the low-delta band, $|\deldelS{}|\leq3.1$ and $|\deldelR{}|\leq3.8$. Qwen3-8B carries direct call-stream evidence of compensation. R1-Distill-14B is shown at its screening pair and mid-session diagnostic pair.}\label{fig:signature}
\end{figure*}

Among the factors this study varied, model identity is the clearest
differentiator of the pooled response. Parameter size does not order the
responses, dose invariance is established only for the dosed model, and
the task subset modulates magnitude and can reverse a near-zero sign.
These findings give Definition~\ref{def:property} its operational footing:
a model version carries a measurable reading under a fixed protocol, not a
universal response law across every task mixture and deployment.
Figure~\ref{fig:signature} locates the model pairs on both indicators and
expands the low-response cluster.

\subsection{Three behavioral signatures}
\label{sec:res53}

Deletion leaves three behavioral signatures, compensation, looping, and
quiet failure, together with a low-delta reference for models whose two
screening indicators remain small. The groups of Section~\ref{sec:res51}
partition outcomes on the success axis, whereas the signatures describe
behavior on a named condition pair. For any ordered pair of conditions,
write
\begin{equation}
\Delta_R(C_a\!\rightarrow\!C_b)=R(C_b)-R(C_a),
\label{eq:pairrep}
\end{equation}
so that $\deldelR=\Delta_R(C1\!\rightarrow\!C2)$. The screening pair is
retention against cross-turn deletion, and a diagnostic pair is chosen
inside a model to expose a distinct behavioral response: mid-session
against retention for R1-Distill-14B, the model whose two readings
separate on the behavior axis. The low-delta reference occupies
$|\deldelS|\le 3.1$ and $|\deldelR|\le 3.8$ on the screening pair, an
operational band for this dataset rather than a universal threshold. A
model is looping when some condition pair involving deletion satisfies
$\Delta_R\ge +30$ points. A model shows quiet failure when
$\deldelS\ge +15$ and $\deldelR< +10$ on the screening pair; the second
clause keeps a model with both readings high out of two signatures at
once. Compensation requires direct call-stream evidence rather than a
small numerical delta alone. This paper applies it to Qwen3-8B, which
rebuilds deleted steps into 410 calls from 6 in the matched session. The
other members of the small-response group remain in the low-delta
reference, and Section~\ref{sec:ana62} distinguishes that numerical
stability from demonstrated re-derivation.

The repetition axis separates the behavioral signatures from their low-delta reference.
Table~\ref{tbl:signature} reports the repetition rate under the three
conditions and Figure~\ref{fig:signature} places each model on the two
indicators. Mid-session deletion raises the repetition rate of
R1-Distill-14B from 16.8 to 66.4 percent, cross-turn deletion raises it
to 22.7 percent, the session re-issues the
same call to the same tool with the same arguments, and the success
axis reads a few points while the behavior axis reads a collapse.
Baichuan-M2-32B falls from 51.8 to 0.5 percent success while its
repetition rate rises to 26.5 percent, GLM-Z1-9B-0414 repeats the shape
at a smaller scale, 19.2 to 2.7 percent with the repetition rate down
7.2 points, both quiet readings, the model gives up without a large repetition rise,
and GLM-Z1-32B holds success within 1.3 points and repetition
within 0.1 points under cross-turn deletion, so a near-zero model keeps
both axes level at once.

The collapsing group expresses quiet failure, while looping appears in
the recovering group. The small-response group divides between the
low-delta reference and the directly evidenced compensation of Qwen3-8B.
R1-Distill-14B is especially informative: its success improves while its
mid-session behavior enters the heaviest looping regime, so the success
axis and the behavior axis carry different facts about the same deletion.
Section~\ref{sec:ana62} interprets that dissociation.

\begin{table}[t]
\caption{Session repetition rate in percent across three conditions, pooled across three seeds. A repeated session issues at least one identical call twice.}\label{tbl:signature}
\footnotesize
\begin{tabular}{@{}lrrrc@{}}
\toprule
Model & Ret. & Cross & Mid & Signature \\
\midrule
Qwen3-8B & 44.4 & 46.1 & 37.0 & compensating \\
Qwen3-14B & 37.1 & 36.4 & 36.6 & low-delta \\
Qwen3-32B & 34.3 & 35.7 & 34.9 & low-delta \\
GLM-Z1-32B & 30.3 & 30.2 & 29.7 & low-delta \\
Ling-mini-2.0 & 38.5 & 34.7 & 36.8 & low-delta \\
R1-Distill-14B & 16.8 & 22.7 & 66.4 & looping \\
GLM-Z1-9B-0414 & 15.0 & 7.8 & 11.2 & quiet \\
Baichuan-M2-32B & 18.5 & 26.5 & 22.0 & quiet \\
\bottomrule
\end{tabular}
\end{table}

\section{Analysis}
\subsection{Same task, three conditions}
\label{sec:ana61}
Table~\ref{tbl:calls} contrasts the actual call streams of matched tasks
under the three conditions. The columns hold what each session actually
did, the contrast carries its own ground truth, and the signatures read
off the difference. The compensating model repeats its retention plan
with extra re-derivation calls, on one task the call count rises from 6
to 410 with the success held, and on another the same plan re-derives
with a single extra call, so the re-derivation cost ranges from one call
to a full re-execution and the outcome stays the same. The looping model
re-issues one call to the end of its budget, and the quiet model ends
in wrong state with a plain log.

\begin{table}[t]
\caption{Matched task sessions across three conditions, with excerpts from the recorded tool-call streams.}\label{tbl:calls}
\setlength{\tabcolsep}{3pt}
\footnotesize
\begin{tabular}{@{}p{0.20\columnwidth}p{0.24\columnwidth}p{0.24\columnwidth}p{0.24\columnwidth}@{}}
\toprule
Model and task & Retention & Cross-turn & Mid-session \\
\midrule
Qwen3-8B, base 128 & plan in 6 calls, done & same plan re-derived, 410 calls, done & plan held, 9 calls, done \\
R1-Distill-14B, miss\_param 17 & reads the state, answers & searches, 22.4\% repeat & \texttt{ls(a=True)} issued 418 times \\
Baichuan-M2-32B, miss\_param 26 & 4 calls, correct state & \texttt{cd(temp)}, one call, silence & \texttt{cd(temp)}, stall \\
\bottomrule
\end{tabular}
\end{table}

\subsection{What the deleted history was doing}
\label{sec:ana62}

A trace of what the compensating model does instead is visible in the
call stream. The cross-turn cells of Qwen3-8B issue more calls than
their retention cells and the success holds, so the work the deleted
trace was doing reappears as extra calls in the stream, and the reading
establishes re-derivation for this model.

For Qwen3-8B the call-count rise establishes re-derivation and prices
its cost. For the dosed Qwen3-32B model, success recovery holds at every
dose. The remaining small-response models establish numerical stability
rather than a particular recovery mechanism: distinguishing re-derivation
from idle trace cargo requires comparing call content within the matched
retention and deletion sessions.

Two roles organize the interpretation of the reasoning history, and the
signatures supply the behavioral evidence. If the history worked as content memory, a store of intermediate
results worth retrieving, the cost of losing it would scale with the
amount deleted, and the dose reading contradicts the scaling, the cost
is absent at every dose for the dosed model. If the history
worked as a continuity scaffold, a structure that keeps the model on its
own track, losing it would break the thread rather than erase a fact,
and the failure would look like disorientation. The looping signature
is a direct observation of that break, the model re-issues calls whose
record vanished with the trace, and the quiet signature is the same
break in a
model that gives up instead of repeating. The observed failures are consistent with a break in continuity, and the
two collapse forms differ by what the model does at that break.
The mid-session arm locates the damage at the break point, R1-Distill-14B
loops at 66.4 percent under mid-session deletion against 22.7 under
cross-turn deletion, so the switch itself throws this model into
repetition harder than deleting the whole history from the start, and
Baichuan-M2-32B loses 37.1 points to deleting the first 40 percent of its
history against 51.3 for all of it, so the damage concentrates in the
early turns. The break point is where the damage lives for sensitive
models, and it is where a routing decision does its harm.

Deletion improves the success of R1-Distill-14B by 3.9 points and at
the same time throws it into the heaviest mid-session looping of the
study. The two readings coexist on different axes, the success axis
reads a burden removed, and the behavior axis reads a thread broken.
The model was carrying a trace it could ill afford, and losing the
trace still breaks its continuity. The three 32B models spread 51.5
points under the same operation, directing the mechanism question past
parameter count toward training lineage. The training-time observations
of Section~\ref{sec:rw-training} motivate that direction, while the
present measurements establish the version-level differentiation on which
it rests.

\subsection{What transfers across task families}
\label{sec:ana-transfer}
The reading of this paper is established on BFCL-class function-calling
tasks, and a task family of another kind re-runs the two cells on its
own traffic, which the two-cell screen of Section~\ref{sec:deploy} makes
cheap. The property of Definition~\ref{def:property} holds inside the
fixed task family, scorer, and sampling protocol, and the transfer
across families sits inside the operational definition the paper
states.

\subsection{What the official checker measures}
\label{sec:ana-checker}
The scorer agrees with the deployed scoring of this project cell by
cell, and the residual exposure is the blind spot of a single checker
family. Terminal-state grading with call order and no-call turns covers
the success axis of this task family, and the repetition channel covers
the behavior axis from the raw call log, so the two indicators read
from independent evidence. Failure modes that leave the terminal state
correct while the path degrades would pass the success axis, and the
behavior channel is the witness for exactly those modes.

\section{Implications for deployment}
\label{sec:deploy}

The screen carries the reading of this paper into a deployment. It runs
one model on the retention cell and the cross-turn cell,
200 tasks each, one seed, in one day on two devices, and reads
\deldelS{} with its interval. The decision has three mutually exclusive
tiers. A reading whose lower bound exceeds $+10$ keeps or pins the
session. A reading whose upper bound falls below $+10$ makes the model
deletion-eligible with log monitoring under Table~\ref{tbl:routing}. A
reading whose interval holds $+10$ also deletes and monitors, but this
tier records threshold uncertainty rather than evidence of insensitivity. The
decision rule rests on a closed-form standard error,
\begin{equation}
\mathrm{SE}(\hat{\Delta})=100\sqrt{\hat{p}_1(1-\hat{p}_1)/n_1+
\hat{p}_2(1-\hat{p}_2)/n_2},
\label{eq:se}
\end{equation}
where $\hat{p}_i=S(C_i)/100$ and $n_i$ is the count of valid sessions in
cell $i$, with cell 1 the retention cell and cell 2 the deletion cell.
The form treats the two cells as independent binomials. For a paired
difference it overstates the variance where the two cells co-vary
positively and understates it where they co-vary negatively, so the screen
reads the
closed form as an order-of-magnitude guide for the width of a cell. Its
half-width $1.96\,\mathrm{SE}(\hat{\Delta})$ is about 9.6 points at this
scale, so calling keep requires a point estimate above roughly 20
points. Three worked readings fix the arithmetic. A model at zero reads
an interval that straddles zero and deletes. A model at $+15$ reads
$[+5.4, +24.6]$, which excludes zero and still holds $+10$ inside, so
the screen places it in the uncertain tier, and the probability that a
truly $+15$ model is called keep under a normal approximation is
$1-\Phi\big((10+1.96\,\mathrm{SE}-\Delta)/\mathrm{SE}\big)$ at
$\Delta=15$, about 17 percent, so a truly $+15$ model usually remains
deletion-eligible even though its positive success loss is not evidence of
insensitivity. A model at
$+25$ reads $[+15.4, +34.6]$ and keeps. The repetition channel rides
along from the call log and supplies a complementary behavioral signal
across the calibrated cells: the looping model rises 49.6 points
in repetition at its largest condition pair while low-delta models
stay within 3.8 on the screening pair, and it reads only
$+5.9$ on the screening pair itself, so the monitor rule of
Table~\ref{tbl:routing} watches the log rather than trusting the
screening reading alone. The channel
reads from the call log at zero scoring cost. Whether the repetition
rise precedes the
failure inside a session is a separate analysis, and the screen uses the
channel as a corroborating reading, and as an early warning it stays
unclaimed.

Table~\ref{tbl:routing} turns the reading into an action, and the
Evidence column states where each action stands. Every routing action is
a decision derived from the measured \deldelS{} and its interval rather
than a measurement in itself, and the summary arm stays
unmeasured until it runs. Keeping the history has a cost of its own, the retained
request grows with every trace the model writes across turns, and a
deployment that pins a sensitive model to one backend pays that growth
out of its context budget. The choice the table forces is the measured
sensitivity of the model against the measured context cost of its
history, and both sides of the ledger are numbers.

\begin{table}[t]
\caption{Routing actions derived from the screening reading, with the evidence basis for each action.}\label{tbl:routing}
\setlength{\tabcolsep}{3pt}
\footnotesize
\begin{tabular}{@{}p{0.40\columnwidth}p{0.34\columnwidth}p{0.18\columnwidth}@{}}
\toprule
Screening reading & Routing action & Evidence basis \\
\midrule
lower bound exceeds $+10$ & keep or pin session & derived \\
upper bound falls below $+10$ & delete, monitor the log & derived \\
interval holds $+10$ & delete, monitor the log & derived \\
mid-session switch & expect the switch cost & derived \\
summary before delete & summarize, then delete & not yet measured \\
\bottomrule
\end{tabular}
\end{table}

As established in Section~\ref{sec:rw-serving}, the practice belongs to
the aggregation ecosystem rather than to provider documentation, and the
screen and the table above are the risk sheet that documentation does
not carry.

\section{Conclusion}
\label{sec:concl}

This paper measured deletion sensitivity, the task cost of deleting the
reasoning history, in 72 valid experimental cells under three rendering
conditions, a multi-seed protocol, and the official checker. The same
operation separates model versions by direction and magnitude, from
$-3.9$ to $+51.3$ points with both endpoint intervals excluding zero.
Parameter size does not order these responses, the dosed model preserves
success across the full deletion range, and task subsets modulate
magnitude and can reverse a near-zero sign. Actual call streams reveal
three behavioral signatures, compensation, looping, and quiet failure,
together with a low-delta reference; none of the three signatures is
identifiable from the final success score alone. Deletion sensitivity is
therefore an operational property of a model version under a fixed
protocol. The two-cell screen turns that property into prior information
for routing, and the repetition channel supplies the behavioral witness
after deployment, so the cost of stripping a trace becomes a priced line
of the serving budget rather than an invisible loss.

\section*{CRediT authorship contribution statement}
\textbf{Yiran Li}: investigation, data curation, writing, original
draft. \textbf{Hanzhang Jia}: conceptualization, methodology, supervision.
\textbf{Liheng Zeng}:
investigation, validation. \textbf{Zhiyu Yang}: investigation,
validation. \textbf{Yi Gao}: investigation, data curation.

\section*{Declaration of competing interest}
The authors declare that they have no known competing financial interests
or personal relationships that could have appeared to influence the work
reported in this paper.

\section*{Data availability}
The session logs, the grading scripts, and the per-cell scoring records
that support the findings of this study are available from the authors
on reasonable request.

\appendix
\section{Request construction under the three conditions}
\label{app:conditions}

Figure~\ref{fig:conditions} shows the message-level construction of one
session under the three conditions of Eq.~(\ref{eq:conds}). Retention
carries the trace, the actions, and the result of every historical turn.
Cross-turn deletion keeps the actions and the result and drops the trace
of every historical turn. Mid-session deletion switches at turn $K$, the
turns at or before $K$ lose their trace and the turns after $K$ keep it.
The current loop keeps its trace under every condition.

\clearpage
\bibliographystyle{cas-model2-names}
\bibliography{cas-refs}

\end{document}